\documentclass{article}

     \usepackage[final]{neurips_2022}

\usepackage[utf8]{inputenc} 
\usepackage[T1]{fontenc}    
\usepackage{hyperref}       
\usepackage{url}            
\usepackage{booktabs}       
\usepackage{amsfonts}       
\usepackage{nicefrac}       
\usepackage{microtype}      
\usepackage{xcolor}         
\usepackage{amsmath,amssymb,amsfonts}
\usepackage[T1]{fontenc}
\usepackage[utf8]{inputenc}

\usepackage{natbib}
\usepackage{enumitem}
\usepackage{amsmath, amssymb}
\usepackage{algorithm}
\usepackage{algpseudocode}
\usepackage{amsthm}
\usepackage{graphicx}
\usepackage{subcaption}

\usepackage{comment}

\usepackage{graphicx}

\newtheorem{theorem}{Theorem}

\title{Optimizer Dynamics at the Edge of Stability with
 Differential Privacy}

\author{%
  Ayana Hussain \\
  Simon Fraser University \\
  \texttt{ayana\_hussain@sfu.ca} 
  \And
  Ricky Fang \\
  Simon Fraser University \\
  \texttt{tfa24@sfu.ca}
}

\begin{document}

\maketitle

\begin{abstract}
Deep learning models can reveal sensitive information about individual training examples, and while differential privacy (DP) provides guarantees restricting such leakage, it also alters optimization dynamics in poorly understood ways. We study the training dynamics of neural networks under DP by comparing
Gradient Descent (GD), and Adam to their privacy-preserving variants. Prior work shows that these optimizers exhibit distinct stability dynamics: full-batch methods train at the Edge of Stability (EoS), while mini-batch and adaptive methods exhibit analogous edge-of-stability behavior. At these regimes, the training loss and the sharpness--the maximum eigenvalue of the training loss Hessian--exhibit certain characteristic behavior. In DP training, per-example gradient clipping
and Gaussian noise modify the update rule, and it is unclear whether these stability patterns persist. We analyze how clipping and noise change sharpness and loss evolution and show
that while DP generally reduces the sharpness and can prevent optimizers from fully reaching the classical stability thresholds, patterns from EoS and analogous adaptive methods stability regimes persist, with the largest learning rates and largest privacy budgets approaching, and sometimes exceeding, these thresholds. These findings highlight the unpredictability introduced by DP in neural network optimization.
\end{abstract}

    
\begin{figure*}[!ht]
    \centering

    \begin{minipage}{0.32\textwidth}
        \centering
        \includegraphics[width=\linewidth]{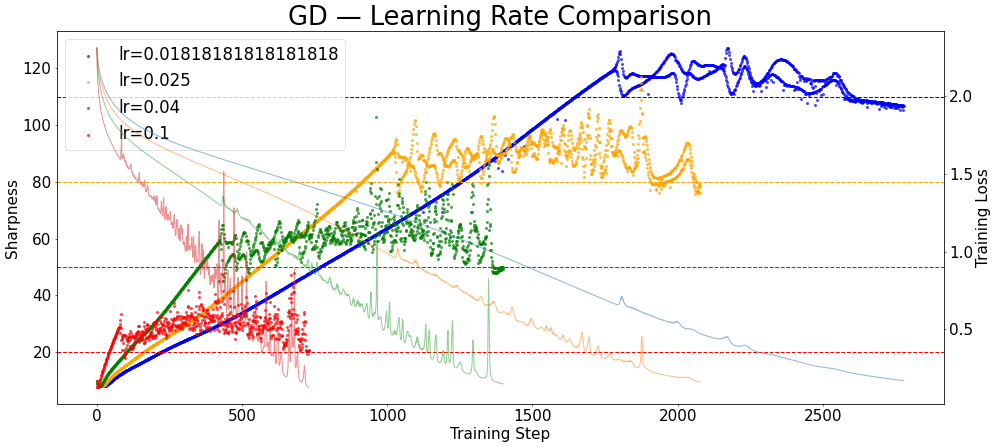}
        \label{fig:sgd-lr}
    \end{minipage}
    \hfill
    \begin{minipage}{0.32\textwidth}
        \centering
        \includegraphics[width=\linewidth]{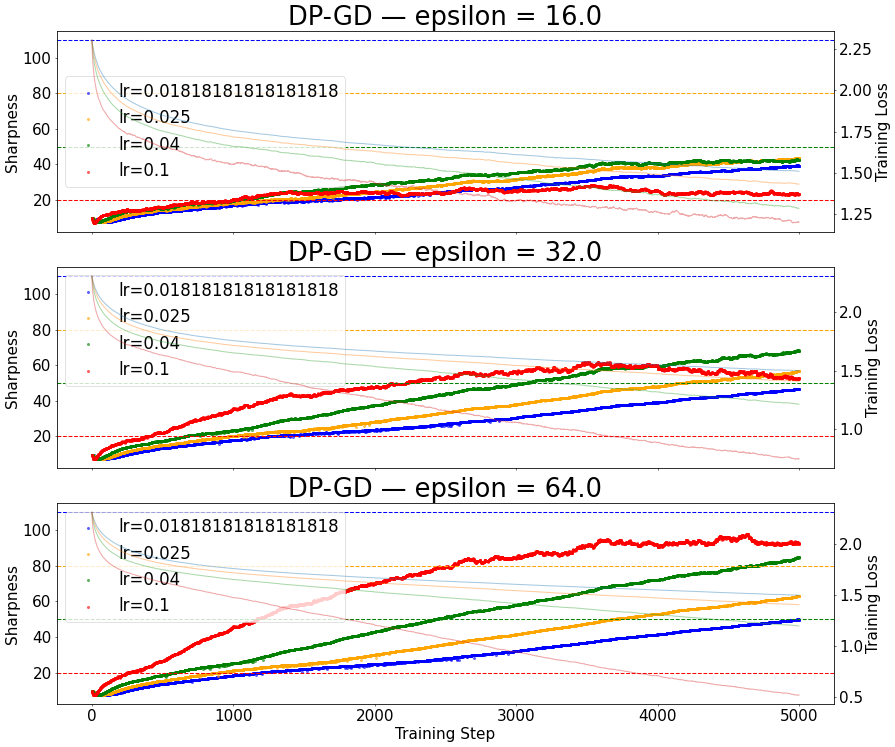}
        \label{fig:sgd-eps}
    \end{minipage}
    \hfill
    \begin{minipage}{0.32\textwidth}
        \centering
        \includegraphics[width=\linewidth]{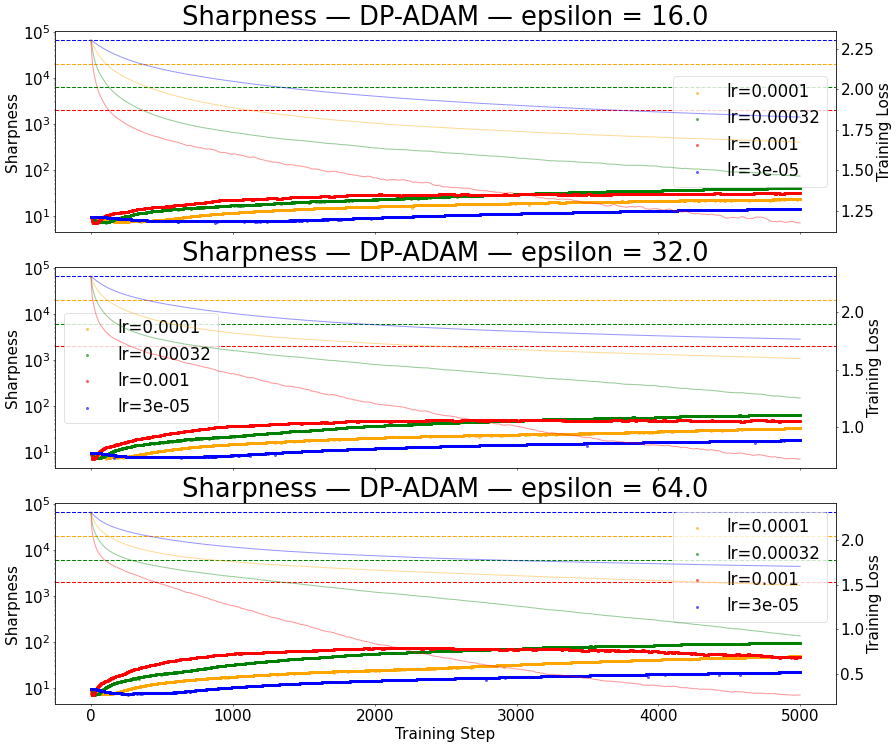}
        \label{fig:adam-sharp-eps}
    \end{minipage}

    \vspace{0.5em}

    \begin{minipage}{0.32\textwidth}
        \centering
        \includegraphics[width=\linewidth]{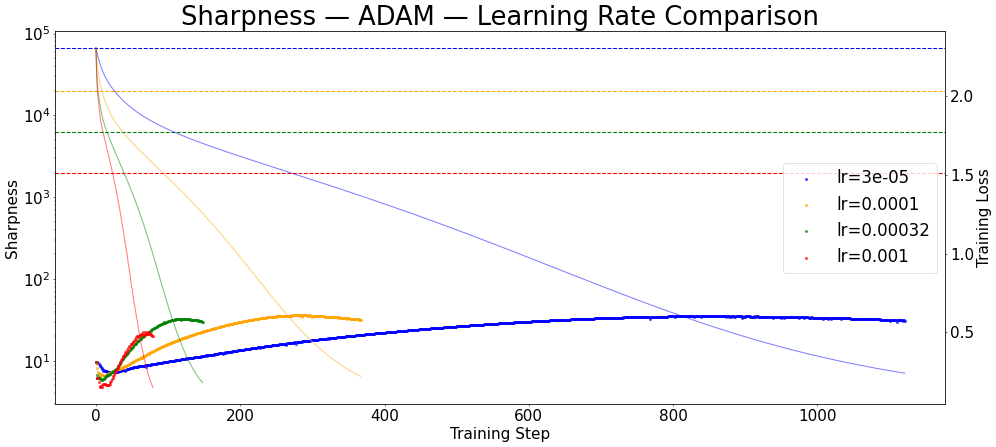}
        \label{fig:adam-sharp-lr}
    \end{minipage}
    \hfill
    \begin{minipage}{0.32\textwidth}
        \centering
        \includegraphics[width=\linewidth]{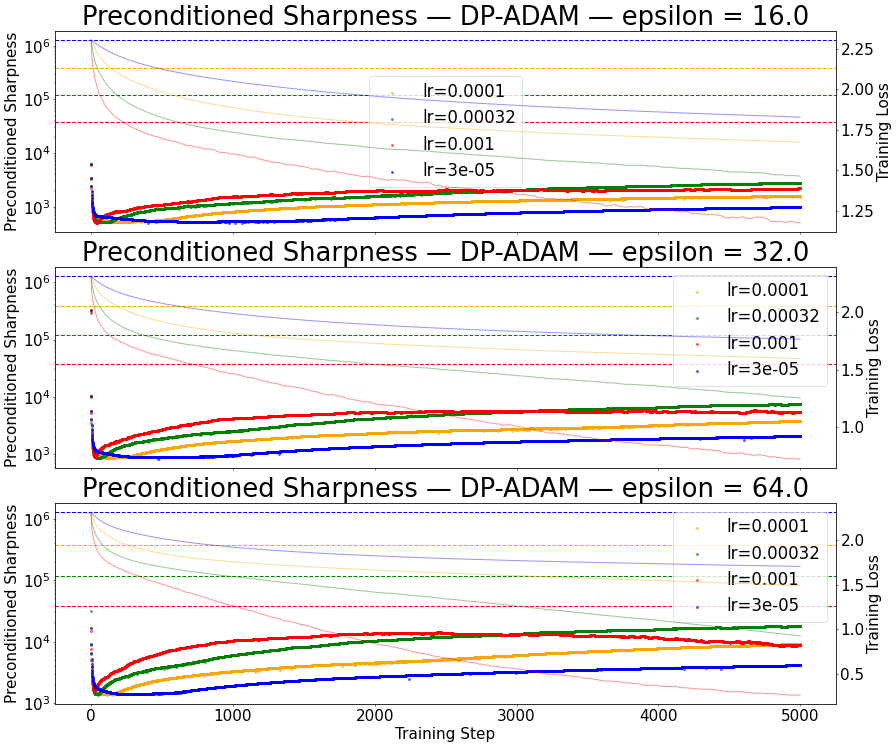}
        \label{fig:adam-pre-eps}
    \end{minipage}
    \hfill
    \begin{minipage}{0.32\textwidth}
        \centering
        \includegraphics[width=\linewidth]{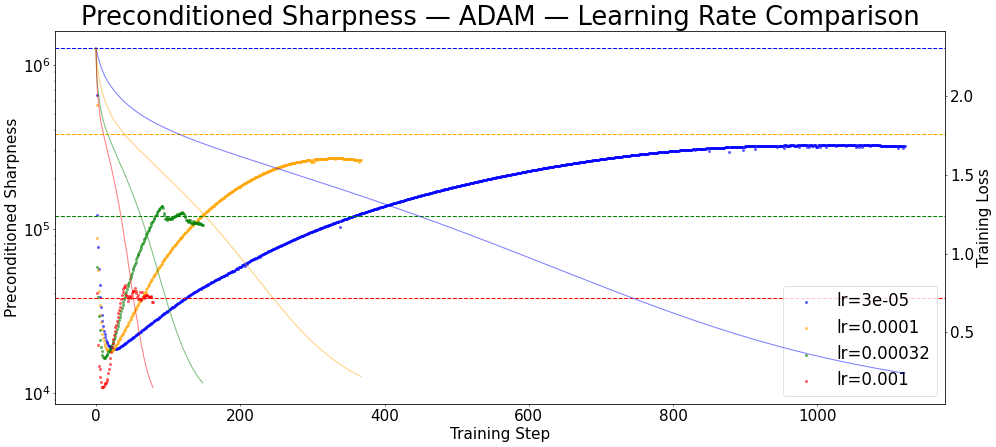}
        \label{fig:adam-pre-lr}
    \end{minipage}

    \caption{\textbf{Sharpness and preconditioned sharpness under full-batch (FB) and DP training.} 
    \textit{Top row:} GD and DP-GD sharpness. For standard GD, sharpness across LR's replicates the Edge of Stability (EoS), with each LR’s stability threshold shown as a solid line in the same color and a lightly colored line for the training loss. Larger LR's reach the threshold faster. Under DP-GD, the largest LR's generally stabilize, particularly for high epsilon, but sharpness often exceeds the normal threshold. 
    \textit{Bottom row and top row right:} Adam and DP-Adam sharpness and preconditioned sharpness. FB Adam exhibits most Adaptive Edge of Stability (AEoS) behavior, however, larger LR's reach the adaptive stability threshold while the smallest one does not. For DP-Adam, sharpness and preconditioned sharpness appear to stabilize together; most LR's and epsilons do not reach the AEoS threshold, although the largest LR and epsilon approach it. These results illustrate how DP modifies sharpness dynamics and underscore the unpredictability of DP training.}
    \label{fig:all-sharpness}
\end{figure*}


\section{Introduction}

Deep learning models are vulnerable to privacy leakage, as model parameters or gradients can reveal information about individual training examples \cite{park2023differentially, wang2021dplis}. Differential privacy (DP) \cite{dwork2006differential} provides a rigorous mathematical guarantee limiting such leakage, and it can be incorporated into many optimization routines by clipping per-example gradients and adding calibrated Gaussian noise to each update \cite{abadi2016deep}. While these modifications protect privacy, they are also known to degrade predictive performance \cite{wang2021dplis} and substantially alter optimization behavior \cite{shi2023make}. Both gradient clipping and noise injection distort the update direction (via induced changes in the model parameters), raising fundamental questions about how DP reshapes the optimization landscape for a broad range of optimizers.

Classical analyses of first-order optimization typically rely on $L$-smoothness assumptions, which bounds curvature along the optimization trajectory (see Appendix A). Convergence guarantees for gradient descent (GD) under this assumption require sharpness--the largest eigenvalue of the training loss Hessian--to remain below $2/\eta$ for learning rate $\eta$ \cite{cohen2021gradient}. Yet modern neural networks routinely violate this condition: as shown by \cite{cohen2021gradient}, sharpness frequently exceeds the smoothness threshold without causing divergence, demonstrating that standard smoothness-based convergence results and analysis of GD through local quadratic Taylor approximations may not reliably predict the behavior of GD when training neural networks.

Stochastic and adaptive optimizers exhibit related behavior. In mini-batch settings, Batch Sharpness--defined as the expected curvature of each mini-batch loss along the direction of its stochastic gradient--becomes the key stabilization property. It increases until it reaches a threshold where updates begin to oscillate, a regime known as the Edge of Stochastic Stability (EoSS) \cite{andreyev2024edge}. Adaptive methods such as Adam reshape the geometry of the problem through a preconditioner (a matrix built from past gradient information that can be thought of as a change of coordinates under which the Hessian is better conditioned), and \cite{cohen2022adaptive} show that Adam operates at an Adaptive Edge of Stability (AEoS), where the preconditioned sharpness---the largest eigenvalue of the Hessian after rescaling by the preconditioner---stabilizes near a deterministic boundary even as raw sharpness continues to grow. These results collectively indicate that GD, SGD, and adaptive methods each exhibit distinct stability regimes that are not captured by classical smoothness assumptions.

Whether these stability patterns persist under DP is not well understood. Because DP imposes per-example clipping and adds noise to gradient updates, any optimizer--GD, SGD, Adam, or others--may experience altered curvature and modified dynamics (see \cite{chua2024private} for discussion). This raises the natural question of how the stability dynamics associated with EoS, EoSS, and AEoS change when training is performed under DP.

In this paper, we conduct an exploratory study to investigate this question using full-batch SGD and Adam in both private and non-private settings, tracking raw and preconditioned sharpness throughout training on CIFAR-10 while varying the LR and privacy budget. We find that DP substantially alters stability behavior across optimizers (see Figure~\ref{fig:all-sharpness}). For GD, strong privacy (large noise) suppresses the growth of sharpness and prevents trajectories from reaching the non-private EoS, while weaker privacy allows sharper regions to be reached, but for small LR's, may still be below the non-private stability thresholds. For Adam, DP causes both raw and preconditioned sharpness to stabilize together at finite values without reaching non-private AEoS thresholds, reverses the usual inverse relationship between LR and sharpness, decreases oscillations, and slows convergence, with sharper solutions for larger LR's and higher privacy budgets. 

Across both methods, the sharpness achieved depends jointly on the LR and privacy parameters, with our results indicating that DP typically produces flatter solutions than non-private training--especially at smaller LR's. These findings are not inconsistent with prior work showing sharper solutions under DP \cite{shi2023make, park2023differentially}; rather, we observe that DP can slow convergence and potentially trap the optimizer in local minima, meaning that with longer training the sharpness could eventually increase to levels comparable to or exceeding non-private runs. Overall, DP induces distinct, and potentially privacy-dependent stability regimes that can give rise to a wide range of unexpected sharpness behaviors compared to the EoS and AEoS dynamics of standard optimizers.

\section{Problem Formulation}

We consider supervised learning on a dataset $\mathcal{D} = \{(x_i, y_i)\}_{i=1}^n$ with $n$ input-output pairs. Let $\theta \in \mathbb{R}^d$ represent the parameters of a neural network. The goal is to minimize the empirical risk $L(\theta) = \frac{1}{n}\sum_{i=1}^n \ell(\theta; x_i, y_i)$, where $\ell(\theta; x_i, y_i)$ is the loss on example $(x_i, y_i)$.

\subsection{Curvature and Stability}

The Hessian matrix $H(\theta) = \nabla^2 L(\theta) \in \mathbb{R}^{d \times d}$ contains second-order partial derivatives of the loss and quantifies local curvature. Sharpness, denoted $\lambda_{\max}(H(\theta))$ or $\lambda_1(H(\theta))$, is the largest eigenvalue of the Hessian and represents how steep the loss surface is in the most sensitive direction (by \cite{cohen2021gradient}).

Gradient descent (GD) updates parameters via $\theta_{t+1} = \theta_t - \eta \nabla L(\theta_t)$, where $\eta > 0$ is the learning rate. For quadratic losses $L(\theta) = \frac{1}{2}\theta^\top A \theta + b^\top \theta + c$ with symmetric $A$, this update is stable only if every eigenvalue satisfies $\lambda_i(A) < \frac{2}{\eta}$; as explained by \cite{cohen2021gradient}, exceeding this causes divergence. 

This motivates the stability threshold $\frac{2}{\eta}$. The breakeven point occurs when sharpness first crosses this threshold. Classical theory may predict divergence beyond this point, yet neural networks continue training successfully in the {Edge of Stability (EoS} regime, where sharpness hovers near $\frac{2}{\eta}$ and loss decreases over long timescales despite short-term oscillations (see \cite{cohen2021gradient}'s paper and section 3 of this paper).

\subsection{Momentum and Adaptive Methods}

\textbf{Momentum} methods are used to accelerate optimizers using momentum and can be parameterized in a "standard" fashion where the momentum vector is $m_{t+1}$, and the momentum update is:  $m_{t+1} = \beta_1 m_t + \nabla L(\theta_t)$ or using exponential moving averages (EMA): $m_{t+1} = \beta_1 m_t + (1-\beta_1)\nabla L(\theta_t)$. Then, the parameter update becomes $\theta_{t+1} = \theta_t - \eta_{t} m_{t}$, where $\beta_1 \in [0,1)$ controls the influence of past gradients. For EMA-style heavy ball momentum for GD, the stability threshold becomes $\frac{2+2\beta_1}{\eta(1-\beta_1)}$ \cite{cohen2022adaptive}.

\textbf{Adam} (as explained by \cite{tang2023dp} maintains first and second moment estimates: $m_t = \beta_1 m_{t-1} + (1-\beta_1)g_t$ and $\nu_t = \beta_2 \nu_{t-1} + (1-\beta_2)g_t^{\circ 2}$, where $g_t$ is the gradient, $\beta_1, \beta_2 \in [0,1)$ are exponential decay rates, and $g_t^{\circ 2}$ denotes element-wise squaring. With bias correction, the estimates are $\hat{m}_t = m_t/(1-\beta_1^t)$ and $\hat{\nu}_t = \nu_t/(1-\beta_2^t)$, and the update is $\theta_{t+1} = \theta_t - \eta_{t} \hat{m}_t/(\sqrt{\hat{\nu}_t} + \epsilon)$, where $\eta > 0$ is the learning rate and $\epsilon > 0$ provides numerical stability. As described by \cite{cohen2022adaptive}, Adam can be viewed as preconditioned GD with dynamic preconditioner $P_{t} = \left[\text{diag}\left(\sqrt{\hat{\nu}_{t}}\right) + \epsilon I\right]$, writing the update as $\theta_{t+1} = \theta_t - \eta_{t} P_{t}^{-1} \hat{m}_{t}$. The relevant stability measure is preconditioned sharpness $\lambda_1(P^{-1}H(\theta))$, which measures curvature after rescaling by the preconditioner (see \cite{cohen2022adaptive}). At the Adaptive Edge of Stability (AEoS), preconditioned sharpness equilibrates near the threshold while raw sharpness $\lambda_1(H(\theta))$ continues to rise.

\subsection{Differential Privacy}

\textbf{Differential Privacy (DP)}, described by \cite{dwork2006differential}, ensures that whether any individual's data is included has negligible effect on the algorithm's output. A randomized mechanism $\mathcal{K}$ (i.e., a randomized algorithm that adds controlled noise to ensure privacy) is $(\varepsilon, \delta)$-DP if for all neighboring datasets $D_1, D_2$ differing by one record and for all measurable sets $S \subseteq \mathrm{Range}(\mathcal{K})$,
\[
\Pr[\mathcal{K}(D_1) \in S] \leq e^\varepsilon \Pr[\mathcal{K}(D_2) \in S] + \delta,
\]
where $\varepsilon \ge 0$ is the privacy budget (smaller values correspond to stronger privacy) and $\delta \ge 0$ is a small failure probability.

\textbf{DP-Optimizers.} Differentially private optimization methods enforce privacy by bounding the sensitivity of per-sample gradients via clipping and injecting calibrated Gaussian noise before parameter updates \cite{abadi2016deep,chua2024private}. For each data point, the per-sample gradient $g_i=\nabla_\theta \ell(\theta;x_i,y_i)$ is clipped to norm $C>0$ as $\tilde g_i = g_i \cdot \min(1, C/\|g_i\|_2)$. For a mini-batch $\mathcal B$ of size $B$, the resulting noised gradient is $\bar g = \frac{1}{B}\sum_{i\in\mathcal B}\tilde g_i + \mathcal N(0,\sigma^2 C^2 I)$, where $\sigma$ is the noise multiplier calibrated to achieve $(\varepsilon,\delta)$-DP (see Appendix B for additional details).

\textbf{DP-SGD and DP-Adam} both apply the same clipping and noise mechanism to per-sample gradients to ensure DP. DP-SGD then performs the standard gradient descent update $\theta_{t+1} = \theta_t - \eta \bar g$ \cite{abadi2016deep}, and in our experiments we use full-batch DP-SGD to enable a direct comparison of sharpness dynamics with EoS, which operates in the full-batch training regime. DP-Adam uses the same privatized gradient within the Adam update in place of the standard (non-private) gradient \cite{wu2025differential}. The full pseudocode for both DP-SGD and DP-Adam is provided in Appendix C.



\subsection{Problem Statement}

We investigate how DP's gradient clipping and noise affect stability dynamics observed in non-private training. We compare full batch SGD, and Adam against their DP variants, tracking sharpness $\lambda_{1}(H(\theta_t))$ and preconditioned sharpness $\lambda_1(P_t^{-1}H(\theta_t))$ throughout training. Our goals are to: (1) determine whether DP preserves or disrupts EoS and AEoS phenomena, (2) identify optimizer and privacy-dependent deviations, and (3) provide practical guidance for hyperparameter selection under privacy constraints. 

\section{Related Work}

Our study connects prior work on stability and convergence behavior in non-private optimization with research examining how DP modifies training dynamics in deep learning.

\subsection{Stability Phenomena in Non-Private Optimization}

Classical convergence theory for GD relies on L-smoothness assumptions that bound curvature throughout optimization. However, \cite{cohen2021gradient}, reveal that neural networks violate these assumptions yet train successfully, operating instead at the EoS where sharpness approaches and stabilizes near $2/\eta$ rather than remaining safely below it. They document progressive sharpening, a phase in which sharpness rises continuously with only brief decreases before returning to growth, culminating in a breakeven point where sharpness first reaches the threshold. Beyond this point, the optimizer exhibits self-regulating behavior where it attempts to increase sharpness but is constrained by incipient instability, resulting in sharpness fluctuating just above $2/\eta$. Training loss becomes non-monotonic over short timescales with occasional spikes, yet consistently decreases over longer horizons. Hyperparameter choices significantly impact these dynamics, with larger step sizes and smaller batches yielding lower equilibrium sharpness values (see \cite{cohen2021gradient} for a more comprehensive analysis).

For mini-batch SGD, \cite{andreyev2024edge} show that the maximum Hessian eigenvalue no longer governs stability. Instead, they introduce Batch Sharpness--the expected curvature of each mini-batch training loss Hessian along the direction of its own stochastic
gradient--which governs the regime in which training operates, known as the EoSS. Empirically, Batch Sharpness rises and stabilizes near $2/\eta$, while the maximum Hessian eigenvalue plateaus at lower values. They identify two types of oscillatory behavior in mini-batch SGD. The first is noise-driven oscillation (Type-1), where stochastic gradient variability causes persistent ``wiggling'' around a point even in flat regions, without leading to instability. The second is curvature-driven oscillation (Type-2), where the interaction of Hessian magnitude and step size leads to repeated overshooting, producing dynamics analogous to classical divergence in full-batch training. Importantly, Batch Sharpness serves as an indicator of Type-2 oscillations: such overshooting (``catapult'') events occur only after Batch Sharpness reaches the critical threshold of $2/\eta$, illustrating the distinction between the two oscillation types and highlighting the role of Batch Sharpness in signaling curvature-driven dynamics. Lastly, they demonstrate that naive noise injection into GD cannot replicate EoSS behavior; only the structured noise inherent to mini-batching produces this regime.

\cite{cohen2022adaptive} extend stability analysis to adaptive optimizers, showing that Adam operates at the AEoS where preconditioned sharpness $\lambda_1(P^{-1}H)$ equilibrates near a threshold while raw sharpness $\lambda_1(H)$ continues rising indefinitely. 
They show that non-adaptive optimizers cannot enter high-curvature regions without destabilizing. However, Adam and other adaptive gradient methods can enter these regions by adjusting their preconditioner. At small LRs, training exhibits a repeating cycle: preconditioned sharpness gradually rises until it crosses the stability threshold, triggering explosive oscillations along certain directions. These large gradients increase the second-moment accumulator and hence the preconditioner, which reduces the preconditioned sharpness below the threshold and temporarily stabilizes the system, after which the cycle repeats. At larger LRs, the cycle proceeds similarly, but instability is resolved differently: both sharpness and preconditioned sharpness drop, indicating that the optimizer has moved to a region where curvature is lower in many directions--a behavior analogous to the ``catapult'' effect observed for non-adaptive optimizers. The final sharpness of the solution is influenced by both the learning rate and the preconditioner decay factor $\beta_2$: smaller values of either parameter tend to produce sharper solutions, while extremely small LRs may prevent preconditioned sharpness from reaching the threshold entirely.

\subsection{Differential Privacy in Deep Learning}

Differential privacy has been widely studied in deep learning, where the predominant approach is to clip gradient's and inject noise into first-order optimization methods following the DP-SGD framework \cite{abadi2016deep} and DP-Adam \cite{li2021large}. These methodologies have become the standard for private training and is supported by major software libraries highlighted by \cite{chua2024private}. Differentially private optimization has been applied across numerous domains, including image classification, generative models, diffusion models, language models, medical imaging, spatial querying, advertising, and recommendation systems listed by \cite{chua2024private}, \cite{mireshghallah2020privacy}. The privacy guarantees of these methods depend on the noise scale, dataset size, batch size, number of gradient update steps, and batch construction procedure as explained by \cite{chua2024private}. 

Formal analyses interpret the private training as issuing bounded per-example queries, with clipping ensuring the required sensitivity constraints, 
and leverage composition guarantees (what happens to the overall privacy guarantee when you apply multiple DP mechanisms (or queries) on the same dataset) \cite{chua2024private} (see Appendix B for formal notation). Recent work has also highlighted that DP substantially alters the geometry of training, where clipping and noise may sometimes prevent the model from updating towards the dominant gradient direction \cite{park2023differentially}, sharpen the loss landscape, or even reduce robustness (see \cite{shi2023make}), while optimizers that promote flatter minima can counteract this effect, as explained by \cite{park2023differentially}. Sharpness-Aware Minimization (SAM) has therefore been explored in private and federated settings to promote flatness and improve stability (see \cite{park2023differentially}, \cite{shi2023make}), though it can introduce additional privacy and computational costs. These results underscore that privacy mechanisms interact closely with sharpness, optimization dynamics, and geometric properties of the loss landscape.

\section{Experiments}

In this section, we empirically investigate how DP affects optimization dynamics and sharpness during neural network training. We first state our central hypothesis and provide an overview of the experimental goals, then describe the experimental and optimization setup. Finally, we present and analyze results for both GD and Adam under private and non-private training.

\subsection{Hypothesis and Overview}

The central hypothesis of this work is that DP alters the curvature and sharpness dynamics encountered along the optimization trajectory, thereby disrupting the characteristic EoS and AEoS patterns observed in non-private training. Specifically, the combination of gradient clipping and injected noise perturbs the update direction at every step, inducing changes in the model parameters and leading the model to traverse different regions of the landscape than by standard optimizers \cite{wang2021dplis}. Prior work frequently reports that these perturbations push DP-SGD toward sharper solutions \cite{shi2023make}, and we therefore expect sharper regions of the landscape to be visited more often than in non-private training. At the same time, we anticipate that the magnitude of sharpness encountered during training will depend strongly on hyperparameters--particularly the learning rate--as smaller LRs can slow escape from high-curvature neighborhoods while larger LRs may interact with DP noise in different ways. Consequently, the added noise and constrained gradients can yield slower or less stable convergence, potentially trapping the model in local minima, and producing sharpness behavior that differ from those observed in the non-private setting.

Our main idea is to directly compare the sharpness and loss dynamics of non-private optimization against their DP counterparts. We focus on two optimizers: 1) GD vs.\ DP-GD, and 2) full-batch Adam vs.\ DP-Adam. We measure both the standard maximum Hessian eigenvalue (sharpness) and the preconditioned sharpness induced by Adam’s diagonal preconditioner, together with the training loss, throughout optimization. We run hyperparameter sweeps over LRs, and privacy budgets to investigate when and how DP disrupts or preserves characteristic behaviors, including progressive sharpening, breakeven point timing, threshold equilibration, and loss behavior.

\subsection{Experimental and Optimization Setup}

All experiments are conducted using a two-hidden-layer fully connected network with 200 units per hidden layer and $\tanh$ activations, taking 3072-dimensional inputs and producing 10-class outputs, following prior work \cite{cohen2021gradient} (see Appendix B for exact architecture). Models are trained using full-batch optimization on a subset of 5{,}000 training examples for a maximum of 5{,}000 epochs with deterministic seeding. We perform learning-rate sweeps separately for GD and Adam. For GD, we use LRs $\eta \in \{2/20,\,2/50,\,2/80,\,2/110\}$ from \cite{cohen2021gradient}, while for Adam we use $\eta \in \{10^{-3},\,3.2\times10^{-4},\,10^{-4},\,3\times10^{-5}\}$ from \cite{cohen2022adaptive}. While the learning rate $\eta$ and privacy budget $\varepsilon$ ideally should be chosen in tandem--a point we return to later--we follow prior work and fix $\eta$ for initial analyses to isolate the effects of DP. For DP runs, we sweep privacy budgets $\varepsilon \in \{16,\,32,\,64\}$ with fixed $\delta = 10^{-5}$ and a single fixed gradient clipping bound $C = 3.0$, chosen based on standard values in previous work \cite{abadi2016deep} and held constant to isolate the effect of privacy noise. Smaller values of $\varepsilon$ were not reported due to prohibitively slow convergence under available computational resources. Similarly, only one clipping bound is used due to compute constraints, and clipping sensitivity analysis is left to future work.

All DP training is implemented using the Opacus library \cite{yousefpour2021opacus} with ghost clipping enabled (explained next) and Rényi Differential Privacy accounting (a method to set strict guarantees and track cumulative privacy loss on the $\epsilon$ budget \cite{moreno2025synthetic}) over the full training horizon (see Appendix B for additional details). In ghost clipping, gradient norms are computed after backpropagation and used to rescale the loss before updating clipped gradients. Essentially, it allows for clipping gradients without needing to materialize every example gradient \cite{kong2023unified}. This approach primarily reduces memory consumptions but requires performing an additional backward pass \cite{beltran2025towards}.  Early stopping is applied based on top-1 training accuracy exceeding 99\%.


At each training step, sharpness is estimated as the dominant eigenvalue of the Hessian using power iteration. For Adam, preconditioned sharpness is measured as the dominant eigenvalue of $P^{-1}H$, where $P$ is the diagonal Adam preconditioner extracted directly from the optimizer state. Power iteration is run with a fixed iteration budget and normalization at each step to ensure numerical stability and consistency across runs. All curvature quantities are logged at every training step. The complete pseudocode for the DP optimizers is provided in Appendix C.

\subsection{GD and DP-GD Results}

\paragraph{GD:} We first replicate the classical EoS behavior using non-private GD with our model. Consistent with prior work, all characteristic EoS properties are observed (see Figure~\ref{fig:all-sharpness}(a)). Sharpness exhibits progressive sharpening with small oscillations throughout training, followed by the emergence of a breakeven point and subsequent instability. The sharpness scales inversely with the learning rate $\eta$, with smaller LRs reaching higher terminal sharpness values. 
The smallest learning rate achieves the highest sharpness near $\lambda \approx 130$, while the largest learning rate stabilizes near $\lambda \approx 40$.

\paragraph{DP-GD:}

We next examine DP-GD (see Figure~\ref{fig:all-sharpness}(b)), beginning with the largest learning rate $\eta = 2/20$. For all privacy budgets $\epsilon \in \{16, 32, 64\}$, sharpness initially increases and then stabilizes at a finite value, indicating progressive sharpening followed by flattening. Larger privacy budgets consistently produce higher stabilized sharpness, while smaller $\epsilon$ values result in earlier flattening and substantially lower sharpness, consistent with stronger noise-induced flattening. If this flattening is interpreted as a DP-modified EoS regime, the breakeven point occurs at the onset of stabilization. 
For all values of $\varepsilon$, the sharpness equilibrates above the non-DP threshold, where $\varepsilon = 16$ yields equilibrium sharpness closest to the non-DP threshold, while $\varepsilon = 64$ exhibits the largest sharpness and the greatest deviation.

For smaller LRs $\eta \in \{2/50, 2/80, 2/110\}$, progressive sharpening is still observed for all $(\eta, \epsilon)$ pairs. Higher privacy budgets again consistently lead to higher sharpness, preserving monotonic scaling with $\epsilon$. Flattening is only observed for $\eta = 2/50$ with $\epsilon = 16$. However, given the high level of noise and the minimal overall increase in sharpness, the stabilizing behavior may simply be due to randomness and may disappear (or continue to progressively sharpen) if trained for more epochs. For the remaining cases, convergence is too slow within 5000 epochs to determine whether a DP-modified threshold would eventually emerge. The corresponding losses are largely monotonic across all DP runs, confirming substantially slower convergence relative to the non-private setting (see Table \ref{tab:results_summary} for a summary of results).

\subsection{Full-Batch Adam and DP-Adam Results}

\paragraph{Adam:} We next reproduce the full-batch non-private Adam experiments to verify presence of AEoS behavior (see Figure's~\ref{fig:all-sharpness}(d, f)). All characteristic AEoS properties are present, including progressive sharpening, the emergence of a breakeven point in the preconditioned sharpness, and the expected inverse scaling with the learning rate (i.e., larger LRs correspond to lower raw and preconditioned sharpness). Across LRs, the raw sharpness continues to grow even after the preconditioned sharpness plateaus. For $\eta = 10^{-3}$ and $\eta = 3.2 \times 10^{-4}$, instability is resolved near the end of training, while for $\eta = 10^{-4}$ and $\eta = 3 \times 10^{-5}$, the preconditioned sharpness does not reach the predicted stability threshold. Both sharpness and preconditioned sharpness are lower and losses smoother than in prior AEoS work, which we attribute to potential differences in hyperparameter and architecture choices.

\paragraph{DP-Adam:} We now analyze DP-Adam using the LR’s given in section 4.2 (see Figure~\ref{fig:all-sharpness}(c, e)). For $\eta = 10^{-3}$, all privacy budgets result in both sharpness and preconditioned sharpness exhibiting progressive sharpening followed by stabilization at finite values. If the flattening of the preconditioned sharpness is interpreted as a DP-modified AEoS regime, the breakeven point occurs at the onset of this plateau. Unlike the non-private case, raw sharpness no longer continues to grow beyond the breakeven point. Instability resolution is still observed, but oscillations are significantly damped. Across all LRs for DP-Adam, none of the $(\eta,\epsilon)$ pairs tested reach the corresponding non-private AEoS thresholds. Larger LRs consistently result in higher sharpness, reversing the scaling behavior observed in the non-private case. Similar to DP-GD, higher $\epsilon$ again produces higher sharpness across all runs. Clear flattening of sharpness is only observed for $\eta = 10^{-3}$; for smaller LRs, convergence is too slow within the fixed 5000-epoch budget to determine whether a DP-modified threshold would emerge. Loss curves for all DP-Adam runs are largely monotonic, further confirming slow convergence under privacy constraints (see Table \ref{tab:results_summary} for a summary of results).

\begin{table}[t]
\centering
\caption{Sharpness dynamics for GD and Adam under DP. 
The key equilibration metric is sharpness (maximum Hessian eigenvalue) for GD and 
preconditioned sharpness (maximum eigenvalue of the preconditioned Hessian) for Adam, 
as in EoS and AEoS analyses. 
Behavior: progressive sharpening (PS), stabilized near non-DP threshold (S$\sim$), 
flattened below (F\textless) or above (F\textgreater) the non-DP threshold.}
\label{tab:results_summary}
\begin{tabular}{llccc}
\toprule
Optimizer & DP & Learning Rate $\eta$ & Privacy Budget $\epsilon$ & Behavior \\
\midrule
GD        & No  & $2/20$--$2/110$     & --      & PS $\rightarrow$ EoS \\
GD        & Yes & $2/20$             & 16      & PS $\rightarrow$ S$\sim$ \\
GD        & Yes & $2/20$             & 32--64  & PS $\rightarrow$ F\textgreater \\
GD        & Yes & $2/50$--$2/110$    & 16--64  & PS (slow) \\
\midrule
Adam      & No  & $10^{-3}$--$3\times10^{-5}$ & -- & PS $\rightarrow$ AEoS \\
Adam      & Yes & $10^{-3}$           & 16--64  & PS $\rightarrow$ F\textless \\
Adam      & Yes & $3.2\times10^{-4}$--$3\times10^{-5}$ & 16--64 & PS (slow) \\
\bottomrule
\end{tabular}
\end{table}

\section{Discussion}

The results reveal that DP reshapes optimization dynamics in ways that differ qualitatively from classical EoS and AEoS behavior.  To clarify the implications, we organize the discussion into three parts: (i) how DP modifies sharpness growth and stability in GD and Adam, (ii) why these findings do not conflict with prior claims of sharper DP solutions, and (iii) what the results suggest for DP training and future work.

\subsection{DP-Induced Modifications to Stability Dynamics}

\paragraph{GD.}  
Sharpness under DP-GD becomes jointly governed by the learning rate and the privacy budget. Specifically, we find that the stabilization point shifts with each $(\eta,\epsilon)$ pair, and the final sharpness may lie above or below the non-DP EoS threshold.  Larger~$\epsilon$ consistently yields higher stabilized sharpness, while for smaller LR's and $\epsilon$ values, sharpness plateaus far below non-private levels, indicating that DP noise can halt curvature growth before reaching the classical breakeven point, although additional analysis would be required to confirm this effect.

\paragraph{Adam.}  
DP similarly modifies Adam’s behavior.  Across all LRs and privacy budgets, both raw and preconditioned sharpness remain far below the non-private AEoS thresholds.  Unlike the non-private case, the two curves plateau together: once the preconditioned sharpness stabilizes, the raw sharpness no longer continues to grow.  This coupled flattening suggests a DP-modified stability regime in which Adam remains in low-curvature regions for extended periods.

\subsection{Relation to Prior Claims About Sharp or Flat Minima}

Prior work presents a mixed picture of how DP shapes the loss landscape.  
Some studies show that DP can obstruct movement toward sharper regions by distorting gradient direction through clipping, while the added noise can impede the convergence of the model weights to the optimum \cite{park2023differentially}, introducing training instability that can leave optimization near local minima \cite{ding2024delving}, or favoring noise-resilient flat areas through smoothing-based objectives \cite{wang2021dplis}. Analyses of DP-SGD demonstrate that injected noise can cause large variability across runs and lead the optimizer to converge to either flatter or sharper minima, with instability increasing when convergence occurs near sharper regions \cite{wang2021dplis}.  
Other work reports settings where DP produces sharper solutions or combines DP with sharpness-aware training to improve performance \cite{shi2023make, park2023differentially}.  
These findings indicate that the effect of DP on sharpness depends strongly on how clipping, noise, and the optimizer interact in a given training setup.

The experiments in our work demonstrate that across both GD and Adam, the sharpness reached by the model depends jointly on the learning rate and the privacy parameters, with smaller step sizes and $\varepsilon$ values under DP often leading to persistently flatter solutions than their non-private counterparts within the same compute budget. In our experiments, larger LRs--particularly when combined with larger privacy budgets (weaker privacy)--still drive the optimizer toward higher-curvature regions. Thus, if DP slows movement toward such regions or holds the optimizer in a flat basin for many epochs, then longer or differently tuned training could still yield the sharp solutions observed elsewhere. In this view, our observations may reflect delayed curvature growth rather than an inherent limitation on DP’s ability to reach sharp minima.

\subsection{Implications for Private Optimization}

In summary, our results indicate that DP induces stability regimes that diverge from standard EoS/AEoS dynamics.  
DP can delay sharpness growth, shift or suppress stability thresholds, or create long-lived flat-plateau phases in both GD and Adam.  
These behaviors complicate optimization and suggest several long-term directions for future work:
(i) developing theory for DP-specific stability dynamics,
(ii) designing optimizers robust to clipping and noise, and
(iii) developing systematic ways for jointly tuning LRs and privacy parameters rather than using non-private heuristics. 

\section{Conclusion and Future Work}

In this paper, we show that DP alters the evolution of sharpness, training loss, and stability thresholds in optimizer dynamics. For DP-GD with large LRs, sharpness occasionally reaches (and may surpass) the classical non-private EoS thresholds, but for smaller LRs it converges much more slowly, suggesting that progressive sharpening may continue over longer training. In DP-Adam, the preconditioned sharpness never reaches the classical non-private AEoS thresholds. In both DP-GD and DP-Adam, higher noise levels consistently reduce sharpness, indicating that DP initially biases optimization toward flatter solutions under our training regimes. These findings show that optimizer behavior under noise and clipping cannot be assumed to follow non-private dynamics, and hyperparameter choices such as LRs, optimizer properties, and noise scale must explicitly account for these DP-induced modifications and be tuned in coordination.

Future work will explore longer training to determine whether smaller LRs eventually approach DP-modified thresholds, isolate the separate effects of clipping and noise on sharpness and stability, and test scaling to larger models and datasets. Additional studies will investigate batch size effects, alternative clipping bounds, mini-batch DP-SGD, and different momentum types to compare DP-SGD with and without momentum and statically preconditioned DP-GD against DP-Adam. Ultimately, these investigations aim to provide actionable guidance for selecting hyperparameters and optimizer strategies when training private models, ensuring both convergence and stability in practice.

\section{Acknowledgments}
This work was completed as part of the Machine Learning: Optimization course at Simon Fraser University. 
We thank our instructor Dr. Sharan Vaswani for their feedback and guidance throughout multiple iterations of this project.

\bibliographystyle{plainnat}
\bibliography{ref}

@article{cohen2021gradient,
  title={Gradient descent on neural networks typically occurs at the edge of stability},
  author={Cohen, Jeremy M and Kaur, Simran and Li, Yuanzhi and Kolter, J Zico and Talwalkar, Ameet},
  journal={arXiv preprint arXiv:2103.00065},
  year={2021}
}

@article{andreyev2024edge,
  title={Edge of stochastic stability: Revisiting the edge of stability for sgd},
  author={Andreyev, Arseniy and Beneventano, Pierfrancesco},
  journal={arXiv preprint arXiv:2412.20553},
  year={2024}
}

@article{cohen2022adaptive,
  title={Adaptive gradient methods at the edge of stability},
  author={Cohen, Jeremy M and Ghorbani, Behrooz and Krishnan, Shankar and Agarwal, Naman and Medapati, Sourabh and Badura, Michal and Suo, Daniel and Cardoze, David and Nado, Zachary and Dahl, George E and others},
  journal={arXiv preprint arXiv:2207.14484},
  year={2022}
}

@inproceedings{dwork2006differential,
  title={Differential privacy},
  author={Dwork, Cynthia},
  booktitle={International colloquium on automata, languages, and programming},
  pages={1--12},
  year={2006},
  organization={Springer}
}

@article{tang2023dp,
  title={DP-Adam: Correcting DP Bias in Adam's Second Moment Estimation},
  author={Tang, Qiaoyue and L{\'e}cuyer, Mathias},
  journal={arXiv preprint arXiv:2304.11208},
  year={2023}
}

@article{zhao2019reviewing,
  title={Reviewing and improving the Gaussian mechanism for differential privacy},
  author={Zhao, Jun and Wang, Teng and Bai, Tao and Lam, Kwok-Yan and Xu, Zhiying and Shi, Shuyu and Ren, Xuebin and Yang, Xinyu and Liu, Yang and Yu, Han},
  journal={arXiv preprint arXiv:1911.12060},
  year={2019}
}

@article{chua2024private,
  title={How private are dp-sgd implementations?},
  author={Chua, Lynn and Ghazi, Badih and Kamath, Pritish and Kumar, Ravi and Manurangsi, Pasin and Sinha, Amer and Zhang, Chiyuan},
  journal={arXiv preprint arXiv:2403.17673},
  year={2024}
}

@inproceedings{abadi2016deep,
  title={Deep learning with differential privacy},
  author={Abadi, Martin and Chu, Andy and Goodfellow, Ian and McMahan, H Brendan and Mironov, Ilya and Talwar, Kunal and Zhang, Li},
  booktitle={Proceedings of the 2016 ACM SIGSAC conference on computer and communications security},
  pages={308--318},
  year={2016}
}

@article{mireshghallah2020privacy,
  title={Privacy in deep learning: A survey},
  author={Mireshghallah, Fatemehsadat and Taram, Mohammadkazem and Vepakomma, Praneeth and Singh, Abhishek and Raskar, Ramesh and Esmaeilzadeh, Hadi},
  journal={arXiv preprint arXiv:2004.12254},
  year={2020}
}

@inproceedings{shi2023make,
  title={Make landscape flatter in differentially private federated learning},
  author={Shi, Yifan and Liu, Yingqi and Wei, Kang and Shen, Li and Wang, Xueqian and Tao, Dacheng},
  booktitle={Proceedings of the IEEE/CVF conference on computer vision and pattern recognition},
  pages={24552--24562},
  year={2023}
}

@inproceedings{park2023differentially,
  title={Differentially private sharpness-aware training},
  author={Park, Jinseong and Kim, Hoki and Choi, Yujin and Lee, Jaewook},
  booktitle={International Conference on Machine Learning},
  pages={27204--27224},
  year={2023},
  organization={PMLR}
}

@inproceedings{dwork2010differential,
  title={Differential privacy under continual observation},
  author={Dwork, Cynthia and Naor, Moni and Pitassi, Toniann and Rothblum, Guy N},
  booktitle={Proceedings of the forty-second ACM symposium on Theory of computing},
  pages={715--724},
  year={2010}
}

@article{kumar2024gd,
  title={GD doesn't make the cut: Three ways that non-differentiability affects neural network training},
  author={Kumar, Siddharth Krishna},
  journal={arXiv preprint arXiv:2401.08426},
  year={2024}
}

@article{li2021large,
  title={Large language models can be strong differentially private learners},
  author={Li, Xuechen and Tramer, Florian and Liang, Percy and Hashimoto, Tatsunori},
  journal={arXiv preprint arXiv:2110.05679},
  year={2021}
}

@article{tan2016barzilai,
  title={Barzilai-Borwein step size for stochastic gradient descent},
  author={Tan, Conghui and Ma, Shiqian and Dai, Yu-Hong and Qian, Yuqiu},
  journal={Advances in neural information processing systems},
  volume={29},
  year={2016}
}

@article{loshchilov2016sgdr,
  title={Sgdr: Stochastic gradient descent with warm restarts},
  author={Loshchilov, Ilya and Hutter, Frank},
  journal={arXiv preprint arXiv:1608.03983},
  year={2016}
}

@article{ding2024delving,
  title={Delving into differentially private transformer},
  author={Ding, Youlong and Wu, Xueyang and Meng, Yining and Luo, Yonggang and Wang, Hao and Pan, Weike},
  journal={arXiv preprint arXiv:2405.18194},
  year={2024}
}

@article{wang2021dplis,
  title={Dplis: Boosting utility of differentially private deep learning via randomized smoothing},
  author={Wang, Wenxiao and Wang, Tianhao and Wang, Lun and Luo, Nanqing and Zhou, Pan and Song, Dawn and Jia, Ruoxi},
  journal={arXiv preprint arXiv:2103.01496},
  year={2021}
}

@article{beltran2025towards,
  title={Towards Efficient and Scalable Implementation of Differentially Private Deep Learning},
  author={Beltran, Sebastian Rodriguez and Tobaben, Marlon and J{\"a}lk{\"o}, Joonas and Loppi, Niki Andreas and Honkela, Antti},
  year={2025}
}

@article{kong2023unified,
  title={A unified fast gradient clipping framework for DP-SGD},
  author={Kong, Weiwei and Munoz Medina, Andres},
  journal={Advances in Neural Information Processing Systems},
  volume={36},
  pages={52401--52412},
  year={2023}
}

@article{yousefpour2021opacus,
  title={Opacus: User-friendly differential privacy library in PyTorch},
  author={Yousefpour, Ashkan and Shilov, Igor and Sablayrolles, Alexandre and Testuggine, Davide and Prasad, Karthik and Malek, Mani and Nguyen, John and Ghosh, Sayan and Bharadwaj, Akash and Zhao, Jessica and others},
  journal={arXiv preprint arXiv:2109.12298},
  year={2021}
}

@article{moreno2025synthetic,
  title={Synthetic Data Generation and Differential Privacy using Tensor Networks' Matrix Product States (MPS)},
  author={Moreno R, Alejandro and Fentaw, Desale and Palmer, Samuel and Salles de Padua, Ra{\'u}l and Dixit, Ninad and Mugel, Samuel and Or{\'u}s, Roman and Radons, Manuel and Menter, Josef and Abedi, Ali},
  journal={arXiv e-prints},
  pages={arXiv--2508},
  year={2025}
}

@phdthesis{wu2025differential,
  title={Differential privacy in the era of generative AI: promises and challenges},
  author={Wu, Fan},
  year={2025},
  school={University of Illinois Urbana-Champaign}
}

@article{lecuyer2021practical,
  title={Practical privacy filters and odometers with R$\backslash$'enyi differential privacy and applications to differentially private deep learning},
  author={L{\'e}cuyer, Mathias},
  journal={arXiv preprint arXiv:2103.01379},
  year={2021}
}

\appendix

\section{Appendix: Background}

\subsection{Convergence analysis and L-smoothness}
\label{app:lsmoothness}

Many convergence analyses of gradient-based optimization rely on regularity assumptions on the training objective. A commonly used assumption is that the loss function $f(\theta)$ is continuously differentiable and $L$-smooth, meaning that there exists a constant $L$ such that
\begin{equation}
\|\nabla f(x) - \nabla f(y)\| \le L \|x - y\| \quad \forall x,y.
\end{equation}
This condition constrains how rapidly the gradient and curvature of the loss can change and is used to justify guarantees on the decrease of the objective under gradient descent \cite{kumar2024gd}.

Such assumptions are difficult to reconcile with neural network training. Modern loss landscapes are often non-smooth or highly irregular, and smoothness may fail to hold globally. One might attempt to restrict smoothness to a local neighborhood around the optimization trajectory, but this requires training to remain in regions where curvature stays sufficiently controlled.

Empirically, neural networks trained with practical learning rates frequently operate at or near the Edge of Stability, where the maximum eigenvalue of the Hessian approaches or exceeds the threshold implied by smoothness-based convergence guarantees. In this regime, even local $L$-smoothness assumptions are violated, limiting the applicability of classical convergence analyses \cite{cohen2021gradient}.

\section{Appendix: Methods - Additional Explanations}

\subsection{Step-Size Schedules}  
To test whether the resulting sharpness and loss behaviors on DP-GD and DP-Adam were artifacts of a constant learning rate, we evaluated cosine decay schedules \cite{loshchilov2016sgdr}, motivated by the fact that SGD converges under diminishing step sizes \cite{tan2016barzilai} (with the caveat that we run full batch SGD). However, the same flat-plateau dynamics persisted.  This indicates that the altered stability regimes arise primarily from the interaction of clipping, noise, and optimizer dynamics rather than whether the step size is constant or decreasing.

\subsection{Differential Privacy}

\paragraph{Randomized Mechanisms}
A randomized mechanism is any algorithm that uses internal randomness so that its output cannot be predicted deterministically from the input. Differential privacy relies on such randomness to obscure individual contributions, explained by \cite{dwork2006differential}.

\textbf{Example (Laplace-perturbed count).}
Let $Q(D)$ output a count from dataset $D$. A DP mechanism is
\[
M(D) = Q(D) + \text{Lap}(b),
\]
where $\text{Lap}(b)$ is Laplace noise with scale $b$. Even if one record changes, the result distribution changes only slightly, making it difficult to infer any individual’s participation.

\paragraph{Sensitivity ($\Delta$)}
The required amount of noise to add depends on the privacy budget and the sensitivity. Thus, the first step to applying DP is determining the maximum possible change in the output of a computation, known as sensitivity.
The $\ell_2$ sensitivity of a query $Q$ is
\[
\Delta = \max_{D,D' \text{ neighbors}} \|Q(D)-Q(D')\|_2.
\]
The $\ell_2$ Sensitivity, denoted by $\Delta$, is the largest possible change (measured by the $\ell_2$ or Euclidean norm) in the output of the query $Q$ when the input dataset $D$ changes by just one record. $Q$ is the Query or function being computed on the data (e.g., a function that calculates the sum of all ages). $D$ and $D'$ are two Neighboring Datasets, which are two datasets that differ by only a single record (one person added or removed). The symbol $\|\cdot\|_2$ denotes the $\ell_2$ (Euclidean) Norm, a mathematical measure of vector length, used here to quantify the magnitude of the difference between the query outputs.

\paragraph{The Gaussian Mechanism}
To make the output private, the Gaussian Mechanism adds random, normally distributed noise scaled by the sensitivity.
The Gaussian mechanism adds noise \cite{tang2023dp}:
\[
M(D) = Q(D) + \mathcal{N}(0,\sigma^2 C^2I),
\]
The Privacy-Preserving Output is $M(D)$, which is the final result returned to the user, including the True Query Output $Q(D)$ plus the protective random noise $\mathcal{N}(0, \sigma^2C^2 I)$. $\mathcal{N}(0, \sigma^2 C^2 I)$ represents a vector of random numbers drawn from a normal (Gaussian) distribution, where $0$ is the Mean of the noise distribution (meaning the noise is centered at zero), C is the clipping value, and $\sigma^2$ is the Variance of the noise, which is calculated to guarantee privacy. $I$ is the Identity Matrix, which ensures that independent noise is added to every dimension (component) of the query output.

\paragraph{Noise Scales for $(\varepsilon,\delta)$-DP}
The required amount of noise ($\sigma$) depends on the desired privacy guarantees, which are defined by two parameters: $\varepsilon$ and $\delta$.
with two classical noise scales for the Gaussian Mechanism, explained by \cite{zhao2019reviewing}:
\[
\sigma_{\text{Dwork-2006}} = 
\frac{\sqrt{2\ln(2/\delta)}\,\Delta}{\varepsilon}, \qquad
\sigma_{\text{Dwork-2014}} = 
\frac{\sqrt{2\ln(1.25/\delta)}\,\Delta}{\varepsilon}.
\]
The symbol $\sigma$ represents the Standard Deviation of the noise, which is the amount of noise that must be added, calculated to satisfy the $(\varepsilon, \delta)$-DP guarantee. $\varepsilon$ (Epsilon) is the Privacy Budget, the core measure of privacy loss. A smaller $\varepsilon$ means stricter (better) privacy, but requires more noise; it roughly bounds how much more likely an observation is if a person is in the dataset versus not in it. $\delta$ (Delta) is the Failure Probability, a small probability that the privacy guarantee will fail (the privacy budget $\varepsilon$ is violated), typically set to be extremely small (e.g., $10^{-5}$ or less). The term $\ln(\cdot)$ is the Natural Logarithm, a mathematical function used as part of the formula derived to link sensitivity and privacy parameters to the necessary noise level. $\sigma_{\text{Dwork-2006}}$ and $\sigma_{\text{Dwork-2014}}$ are two different mathematical formulas proposed in 2006 and 2014 by Cynthia Dwork for calculating the required noise standard deviation ($\sigma$).
\cite{zhao2019reviewing} explain that the Gaussian mechanism introduced in 2014 adds less noise than the earlier 2006 formulation for parameter regimes in which both mechanisms satisfy $(\varepsilon,\delta)$-DP. However, the original proofs for both mechanisms assume $\varepsilon \le 1$. Subsequent analysis shows that when $\varepsilon$ is large relative to $\delta$, these noise calibrations may fail to provide $(\varepsilon,\delta)$-DP, despite being used in practice \cite{zhao2019reviewing}.

\paragraph{Differential privacy guarantees.}
For completeness, we restate another foundational results from the differential privacy literature \cite{dwork2010differential}. This result formalizes how repeated application of private mechanisms yield quantifiable privacy guarantees.

\begin{theorem}[Sequential composition; \cite{dwork2010differential}]
\label{thm:composition}
Let $K_1$ be an $\varepsilon_1$-differentially private mechanism and $K_2$ be an $\varepsilon_2$-differentially private mechanism. Then the sequential composition $(K_1, K_2)$ satisfies $(\varepsilon_1 + \varepsilon_2)$-differential privacy.
\end{theorem}

\noindent
The composition theorem characterizes how privacy loss accumulates over multiple optimization steps. Although practical DP optimizers typically rely on advanced composition techniques, this theorem provides the conceptual basis for understanding how privacy mechanisms interact with optimization dynamics, which we examine empirically in this work.

\subsection{Rényi Differential Privacy Accounting}

Rényi Differential Privacy (RDP) accounting is an analysis tool for DP composition that provides a principled way to track cumulative privacy loss over multiple computations \cite{lecuyer2021practical}. It expresses privacy guarantees in terms of the Rényi divergence of order $\alpha \in (1, \infty]$ between the distributions of the mechanism's outputs on neighboring datasets. Specifically, a randomized mechanism $M$ satisfies $(\alpha, \epsilon)$-RDP if
\[
D_\alpha(M(D)\|M(D')) \leq \epsilon,
\]
where $D$ and $D'$ differ by a single data point, and 
\[
D_\alpha(P \,\|\, Q) \triangleq \frac{1}{\alpha-1} \log \mathbb{E}_{x \sim Q} \Bigg[ \Big(\frac{P(x)}{Q(x)}\Big)^\alpha \Bigg]
\]
is the Rényi divergence of order $\alpha$.

RDP enjoys an additive composition property: if a sequence of mechanisms each satisfy $(\alpha, \epsilon_i)$-RDP, the combined mechanism satisfies $(\alpha, \sum_i \epsilon_i)$-RDP. This property enables precise tracking of cumulative privacy loss across all training steps and allows the conversion to $(\epsilon, \delta)$-DP: for a given $\delta > 0$, an $(\alpha, \epsilon)$-RDP mechanism also satisfies $(\epsilon + \frac{\log(1/\delta)}{\alpha-1}, \delta)$-DP.  

In practice, Opacus uses RDP accounting to compute the noise multiplier required to achieve a target $(\epsilon, \delta)$-DP budget over the full training horizon. This ensures that our reported privacy guarantees are strict and cumulative across all gradient updates.

\subsection{Network Architecture.}
All experiments use a fully connected feedforward network consisting of an input flattening layer, followed by two hidden layers with 200 units each and $\tanh$ activations, following \cite{cohen2021gradient}. The final linear layer maps the 200-dimensional hidden representation to 10 output classes. Concretely, the architecture is
\[
3072 \;\rightarrow\; 200 \;\rightarrow\; 200 \;\rightarrow\; 10,
\]
with $\tanh$ activations after each hidden layer.

\section{Appendix: Methods - Pseudocode}

In this section, we provide the pseudocode used for the differentially private optimization algorithms considered in our experiments, namely DP-SGD and DP-Adam. For the DP-SGD baseline, we instead implement DP-GD, which follows the same update rule as DP-SGD but omits subsampling. This choice is made to enable a direct comparison with full-batch SGD, ensuring that differences in behavior arise from the optimization dynamics rather than from stochastic minibatch effects.

\subsection{DP-SGD}

\begin{algorithm}[H]
\caption{Differentially Private SGD (DP-SGD) \cite{abadi2016deep}}
\label{alg:dpsgd}
\begin{algorithmic}[1]
\Require Dataset $\{x_i\}_{i=1}^N$, loss $L(\theta)=\frac{1}{N}\sum_i L(\theta,x_i)$, learning rate $\eta_t$, noise scale $\sigma$, sampling probability $q=L/N$, clipping norm $C$, number of iterations $T$
\State Initialize $\theta_0$ randomly
\For{$t=0$ to $T-1$}
    \State Sample batch $L_t$ by including each data point independently with probability $q$
    \For{each $x_i \in L_t$}
        \State $g_t(x_i) \gets \nabla_{\theta_t} L(\theta_t,x_i)$
        \State $\bar g_t(x_i) \gets \frac{g_t(x_i)}{\max\!\left(1,\frac{\|g_t(x_i)\|_2}{C}\right)}$
    \EndFor
    \State $\tilde g_t \gets \frac{1}{|L_t|}\sum_{x_i\in L_t}(\bar g_t(x_i) + \mathcal N(0,\sigma^2 C^2 I))$
    \State $\theta_{t+1} \gets \theta_t - \eta_t \tilde g_t$
\EndFor
\State Compute overall privacy cost $(\varepsilon,\delta)$ using a privacy accountant
\State \Return $\theta_T$
\end{algorithmic}
\end{algorithm}

\subsection{DP-Adam}

\begin{algorithm}[H]
\caption{Differentially Private Adam (DP-Adam) \cite{wu2025differential}}
\label{alg:dpadam}
\begin{algorithmic}[1]
\Require Dataset $D$, loss $\ell(w;x)$, learning rate $\eta$, clipping norm $C$, noise multiplier $\sigma$, batch size $b$, momentum parameters $\beta_1, \beta_2$, stability constant $\gamma$, total steps $T$
\State Initialize $w_0$, $m_0 = 0$, $v_0 = 0$
\For{$t=0$ to $T-1$}
    \State Sample mini-batch $S_t$ using Poisson subsampling
    \For{each $x_i \in S_t$}
        \State $g_i \gets \nabla_{w_t}\ell(w_t;x_i)$
        \State $\tilde g_i \gets g_i / \max\!\left(1,\frac{\|g_i\|_2}{C}\right)$
    \EndFor
    \State $\bar g_t \gets \frac{1}{|S_t|}(\sum_{i\in S_t}\tilde g_i + \mathcal N(0,\sigma^2 C^2 I))$
    \State $m_{t+1} \gets \beta_1 m_t + (1-\beta_1)\bar g_t$
    \State $v_{t+1} \gets \beta_2 v_t + (1-\beta_2)\bar g_t^2$
    \State $\hat m_{t+1} \gets \frac{m_{t+1}}{1-\beta_1^{t+1}}, \quad \hat v_{t+1} \gets \frac{v_{t+1}}{1-\beta_2^{t+1}}$
    \State $w_{t+1} \gets w_t - \eta \frac{\hat m_{t+1}}{\sqrt{\hat v_{t+1}}+\gamma}$
\EndFor
\State \Return $w_T$
\end{algorithmic}
\end{algorithm}

\section{Appendix: Evaluation - Additional Results}

\begin{figure}[htbp!]
    \centering
    \includegraphics[width=0.7\linewidth]{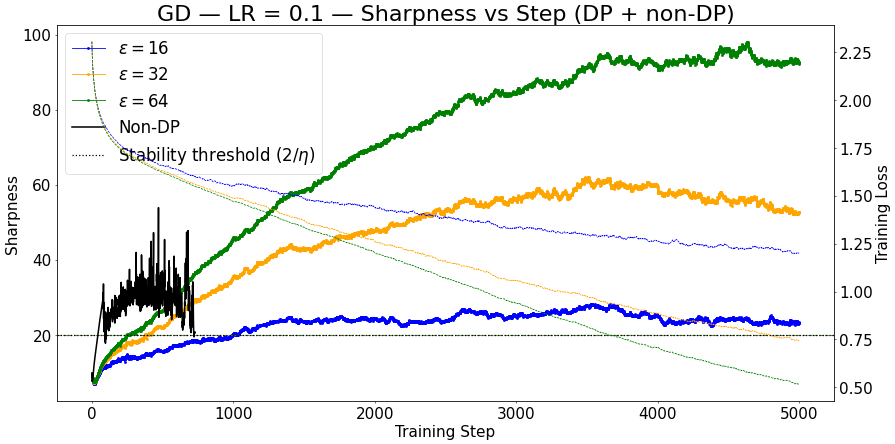}
    \caption{
    \textbf{DP-GD sharpness for the largest learning rate}
    Sharpness trajectories for DP-GD ($\eta=0.1$) across privacy budgets $\varepsilon$.
    Solid, dark curves show sharpness, lighter curves show training loss, and black denotes the non-DP baseline.
    The dotted line marks the stability threshold $2/\eta$.
    Higher $\varepsilon$ yields larger stabilized sharpness, indicating an $\varepsilon$-dependent edge-of-stability regime. If sharpness flattening is interpreted as an edge-of-stability (EoS) behavior, a breakeven point naturally emerges at the point where sharpness stops increasing and begins to stabilize (possibly creating a new DP-induced EoS).
    }
    \label{fig:dp_sgd_sharpness_large}
\end{figure}

\begin{figure}[htbp]
    \centering
    \begin{subfigure}[b]{0.4\linewidth} 
        \centering
        \includegraphics[width=\linewidth]{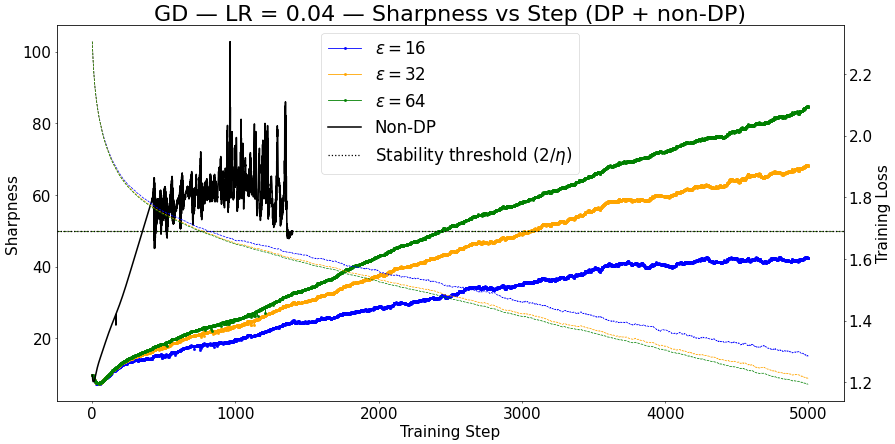}
        \caption{}
        \label{fig:sub1}
    \end{subfigure}
    \hfill
    \begin{subfigure}[b]{0.4\linewidth}
        \centering
        \includegraphics[width=\linewidth]{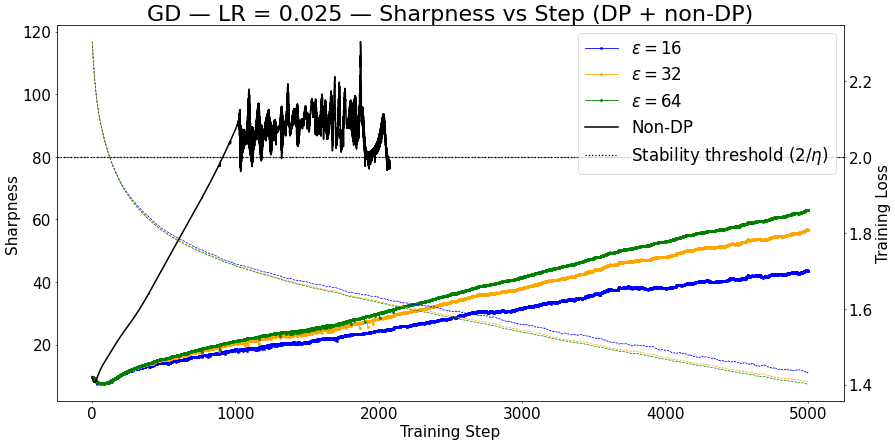}
        \caption{}
        \label{fig:sub2}
    \end{subfigure}
    \hfill
    \begin{subfigure}[b]{0.4\linewidth}
        \centering
        \includegraphics[width=\linewidth]{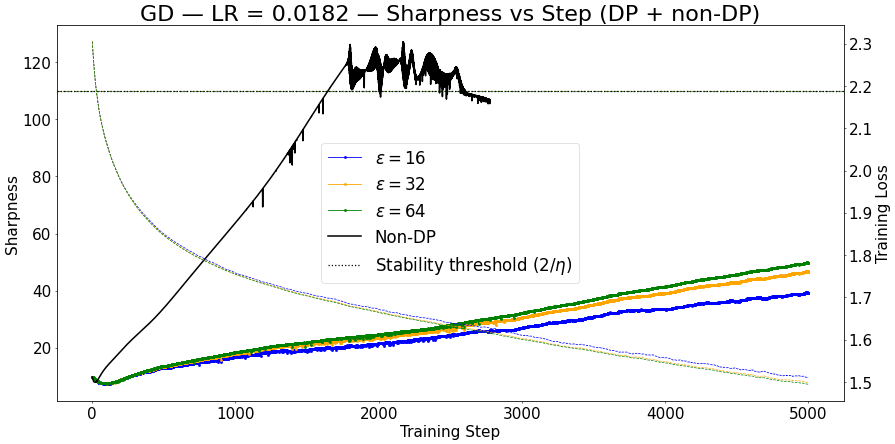}
        \caption{}
        \label{fig:sub3}
    \end{subfigure}
    \caption{
    \textbf{DP-GD sharpness across smaller learning rates.}
    Sharpness dynamics for DP-GD under the same experimental setup as the previous figure, shown for three learning rates: $\eta = 0.04$ (left), $\eta = 0.0182$ (middle), and $\eta = 0.025$ (right).
    Across all learning rate and privacy budget $(\eta,\varepsilon)$ pairs, we observe progressive sharpening, with larger $\varepsilon$ consistently yielding higher sharpness.
    Flattening behavior is partially observed only for $\eta = 2/50$ at $\varepsilon = 16$ (but could be due to randomness); for other configurations, convergence is slower and no definitive edge-of-stability regime can be identified.
    Training losses for DP runs remain largely monotonic, further indicating slow convergence in these settings.
    }

    \label{fig:three_plots_fullwidth}
\end{figure}

\begin{figure}[htbp]
    \centering
    \begin{subfigure}[b]{0.7\textwidth}
        \centering
        \includegraphics[width=\textwidth]{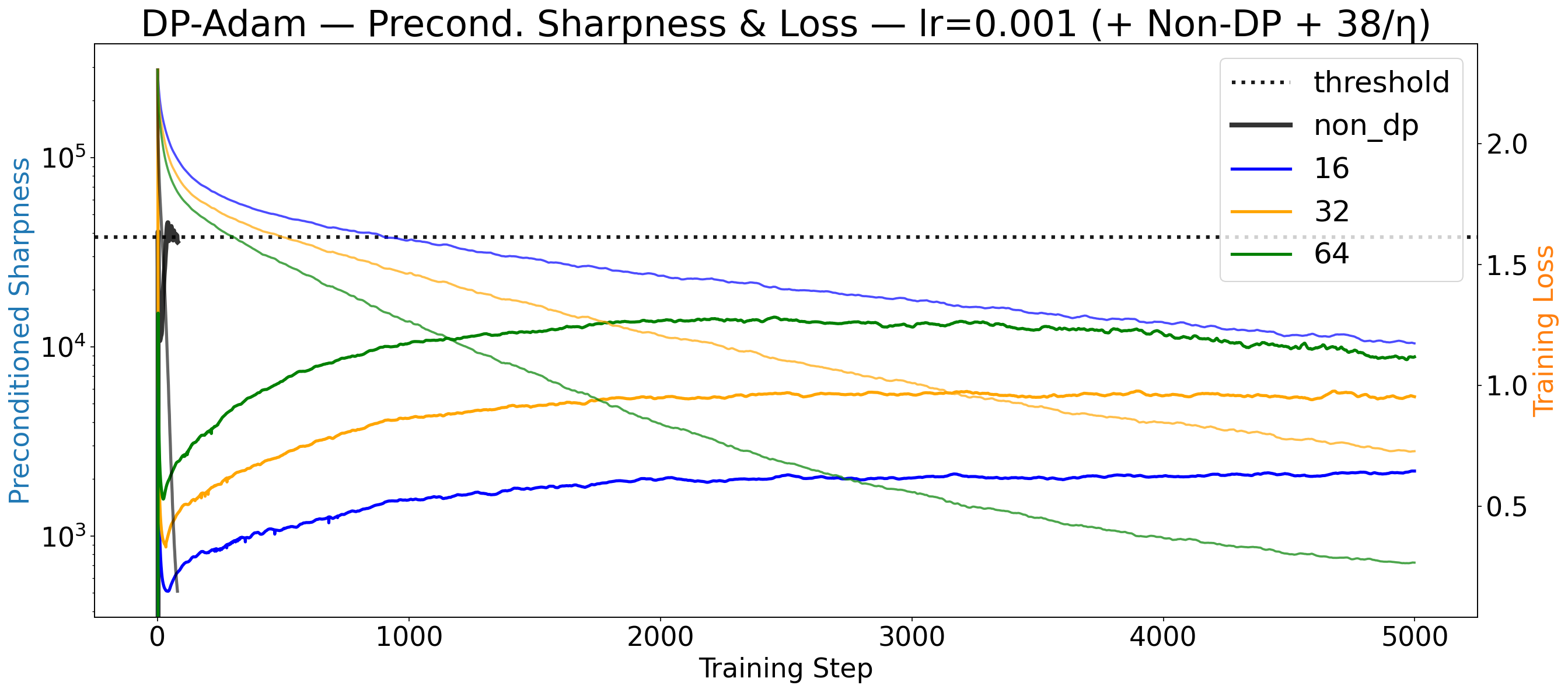}
        \caption{Preconditioned Sharpness vs Training Loss}
        \label{fig:dp_adam_precond_lr1e3}
    \end{subfigure}
    \hfill
    \begin{subfigure}[b]{0.7\textwidth}
        \centering
        \includegraphics[width=\textwidth]{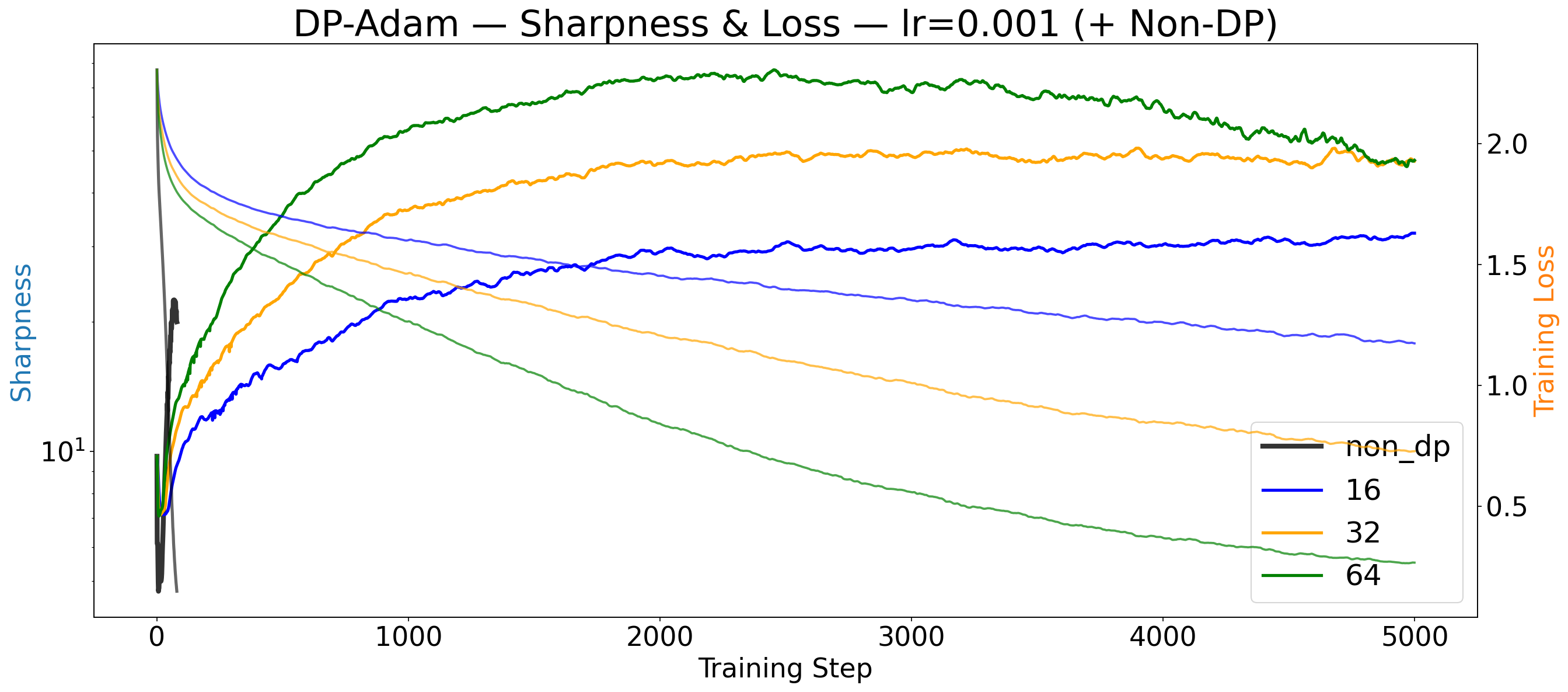}
        \caption{Sharpness vs Training Loss}
        \label{fig:dp_adam_sharp_lr1e3}
    \end{subfigure}
    \caption{
        \textbf{DP-Adam sharpness for the largest learning rate} 
        \textit{Top:} Preconditioned sharpness shows progressive sharpening across $\epsilon \in \{16, 32, 64\}$, with lower $\epsilon$ stabilizing at lower values. The black dotted line marks the Adam stability threshold $38/\eta$; no $\epsilon$ reaches this threshold. The thick black ``Non-DP'' curve (no noise) shows baseline behavior. All loss curves (right $y$-axis) are light solid lines matching their sharpness colors. 
        \textit{Bottom:} Raw sharpness exhibits similar $\epsilon$-dependent flattening. 
        Progressive sharpening persists for preconditioned sharpness but plateaus after breakeven (flattening onset), with reduced oscillatory instability compared to non-private Adam. This suggests a new potential AEoS under DP where breakeven marks stability.
        \label{fig:dp_adam_lr1e3}
    }
\end{figure}

\begin{figure*}[!ht]
    \centering
    \begin{minipage}{0.48\textwidth}
        \centering
        \includegraphics[width=\linewidth]{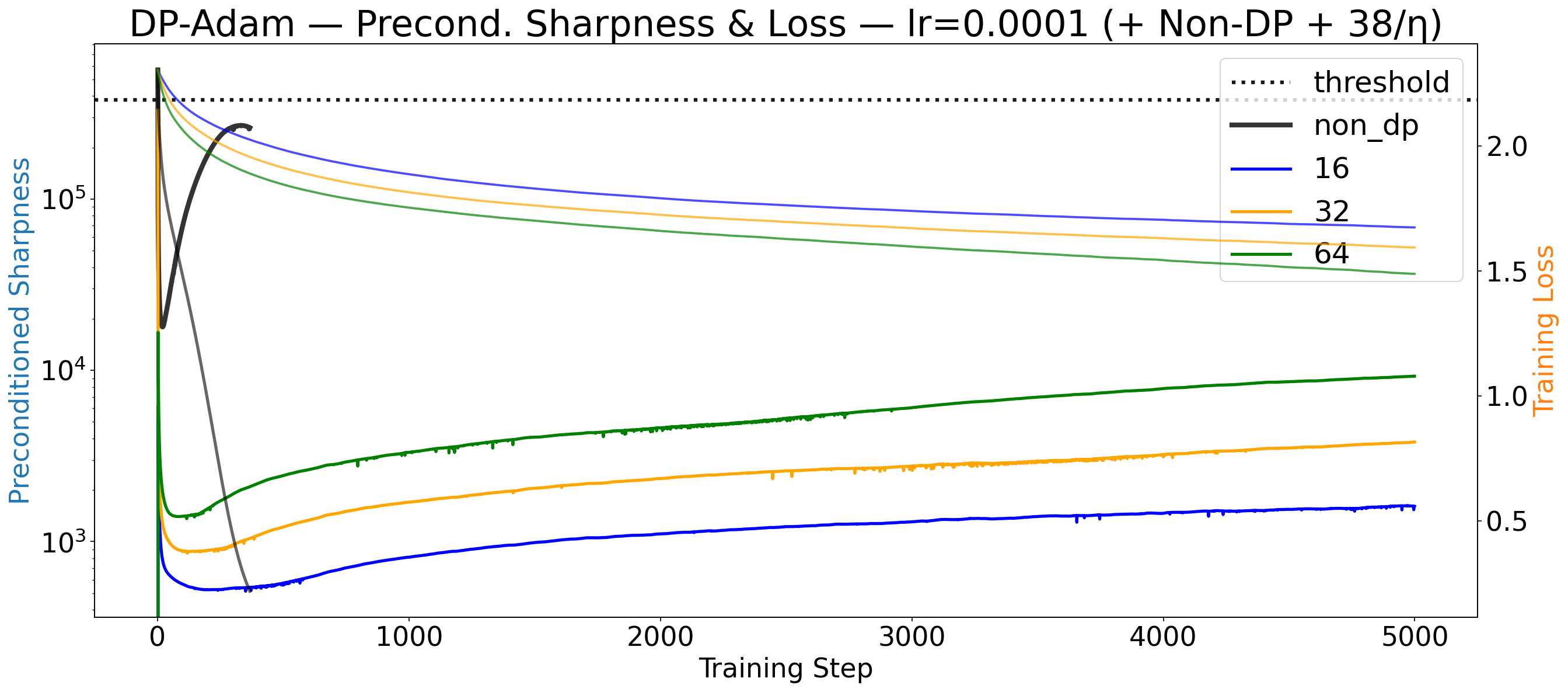}
        \label{fig:row1-col1}
    \end{minipage}
    \hfill
    \begin{minipage}{0.48\textwidth}
        \centering
        \includegraphics[width=\linewidth]{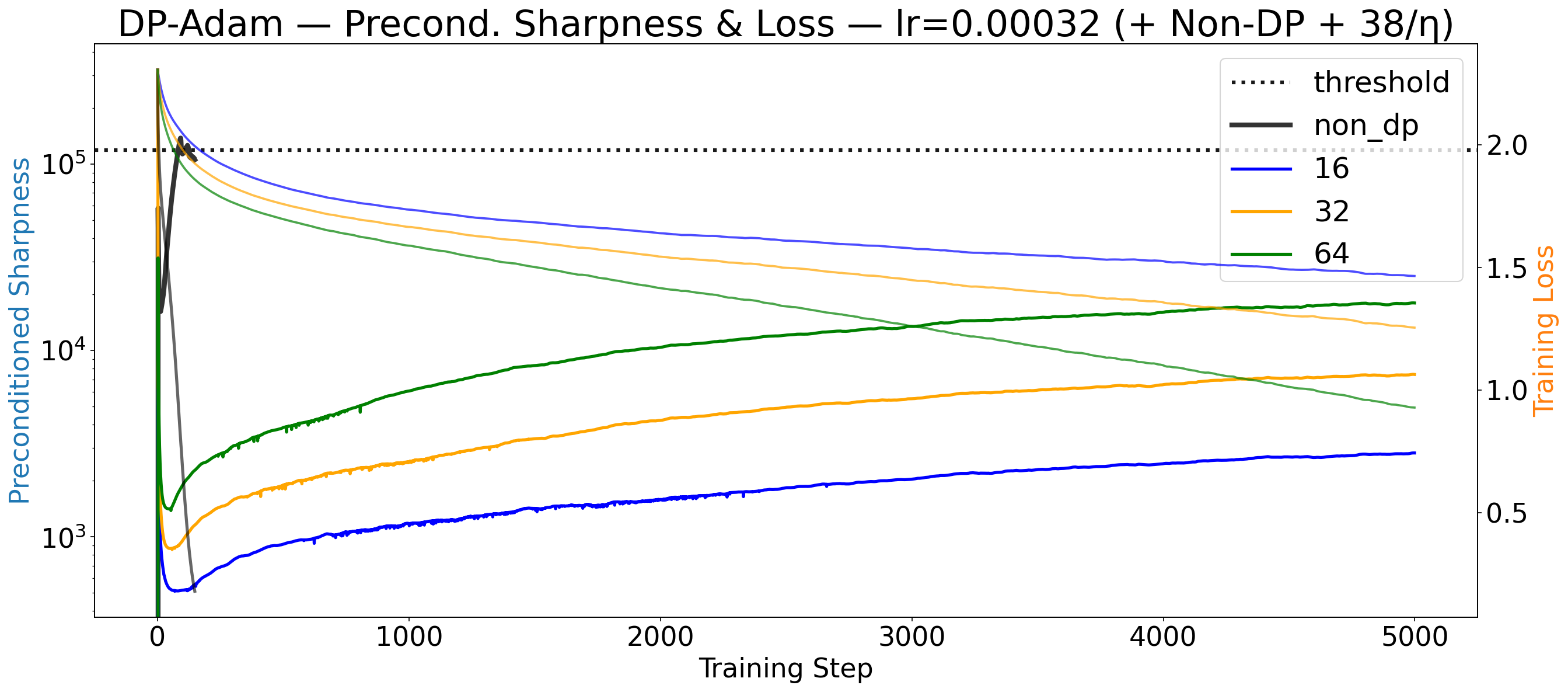}
        \label{fig:row1-col2}
    \end{minipage}

    \vspace{0.6em}

    \begin{minipage}{0.48\textwidth}
        \centering
        \includegraphics[width=\linewidth]{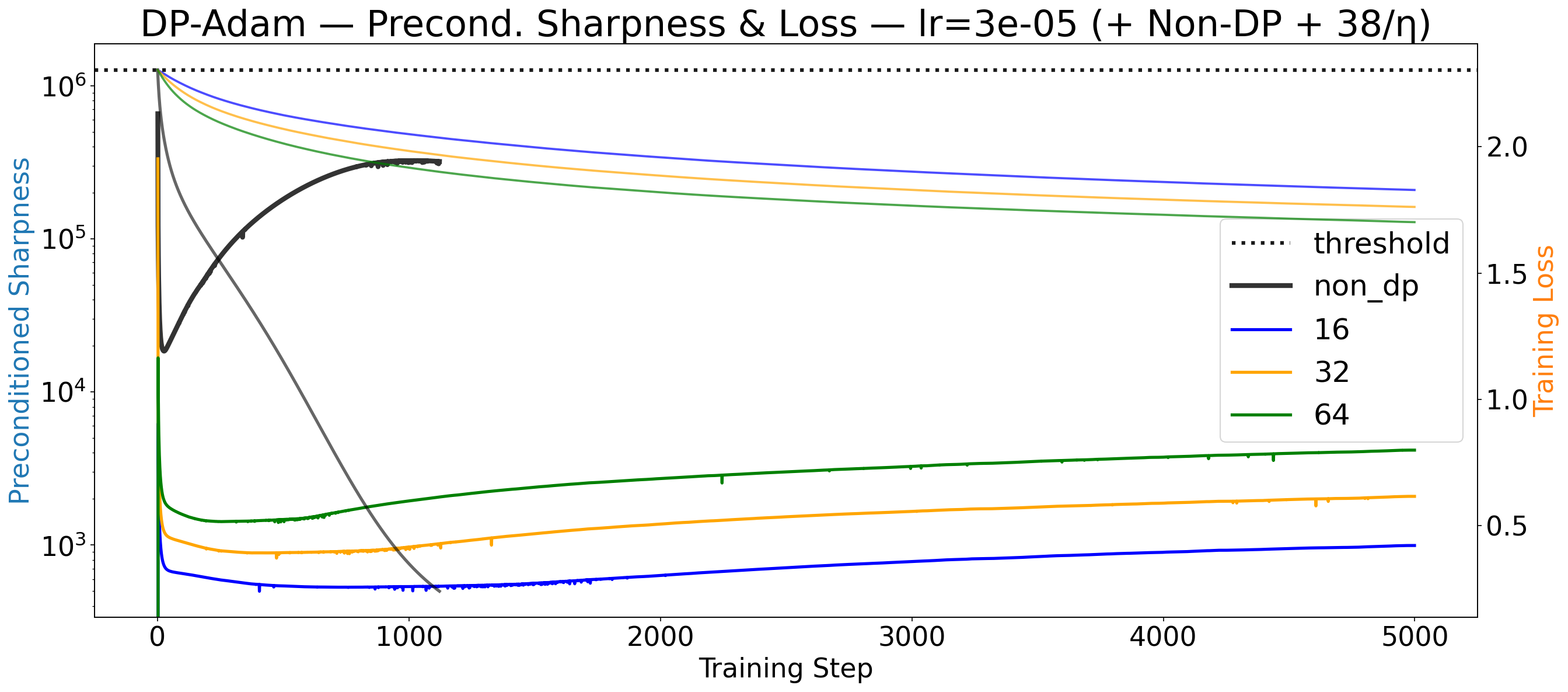}
        \label{fig:row2-col1}
    \end{minipage}
    \hfill
    \begin{minipage}{0.48\textwidth}
        \centering
        \includegraphics[width=\linewidth]{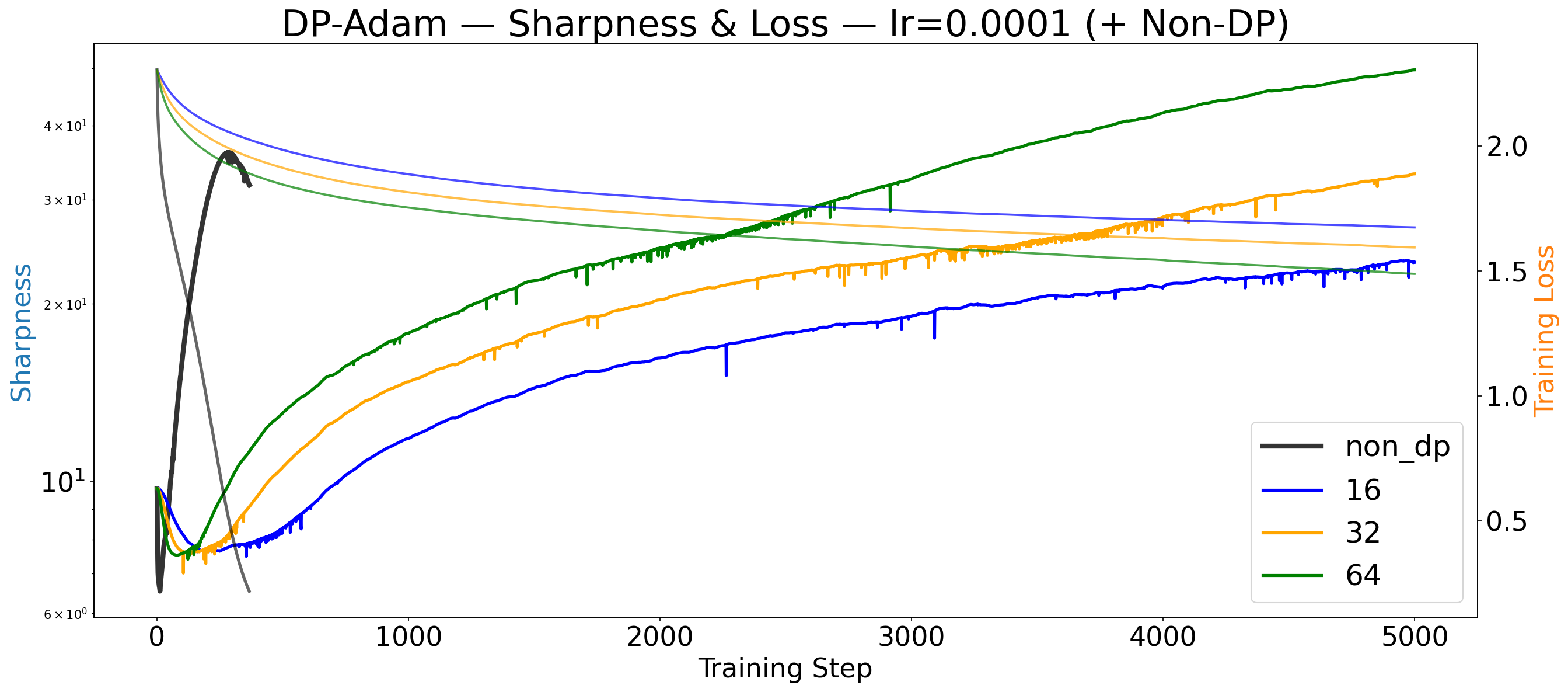}
        \label{fig:row2-col2}
    \end{minipage}

    \vspace{0.6em}

    \begin{minipage}{0.48\textwidth}
        \centering
        \includegraphics[width=\linewidth]{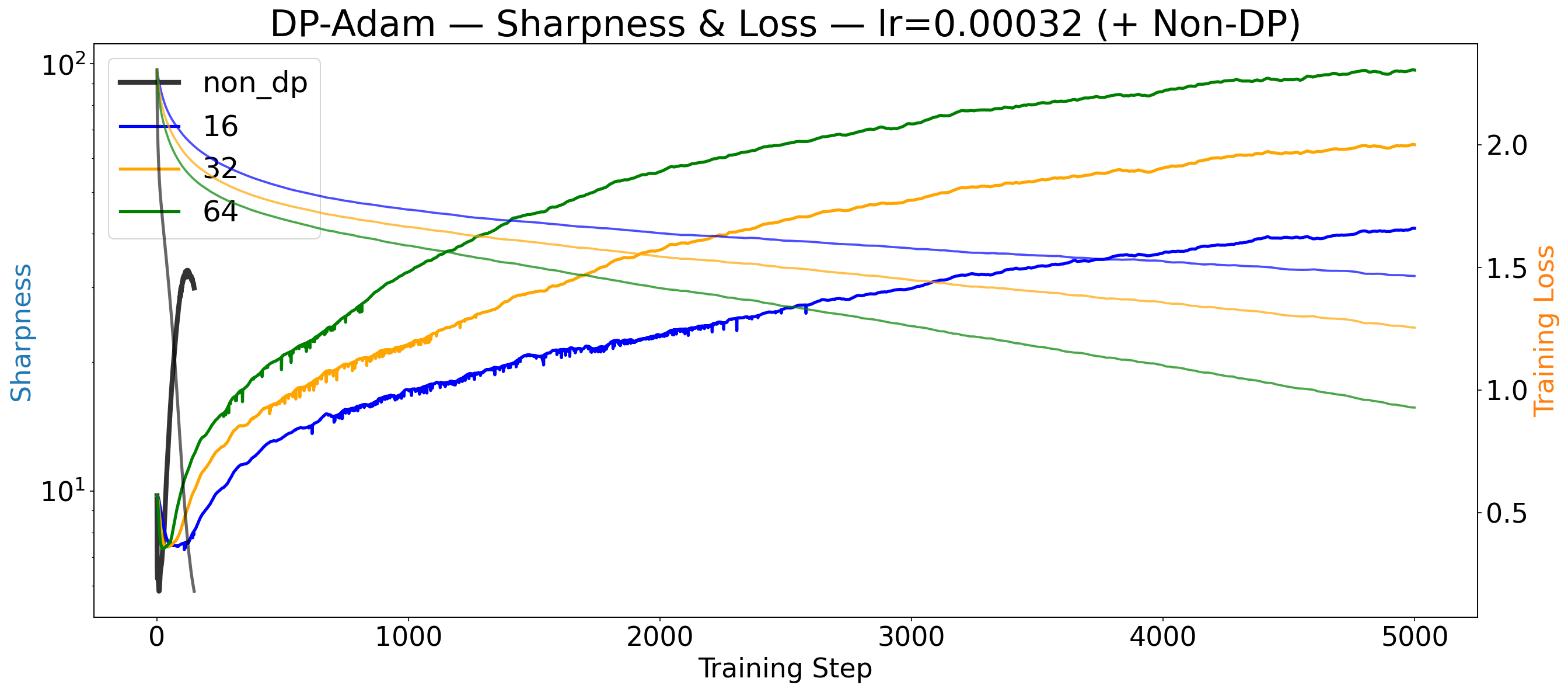}
        \label{fig:row3-col1}
    \end{minipage}
    \hfill
    \begin{minipage}{0.48\textwidth}
        \centering
        \includegraphics[width=\linewidth]{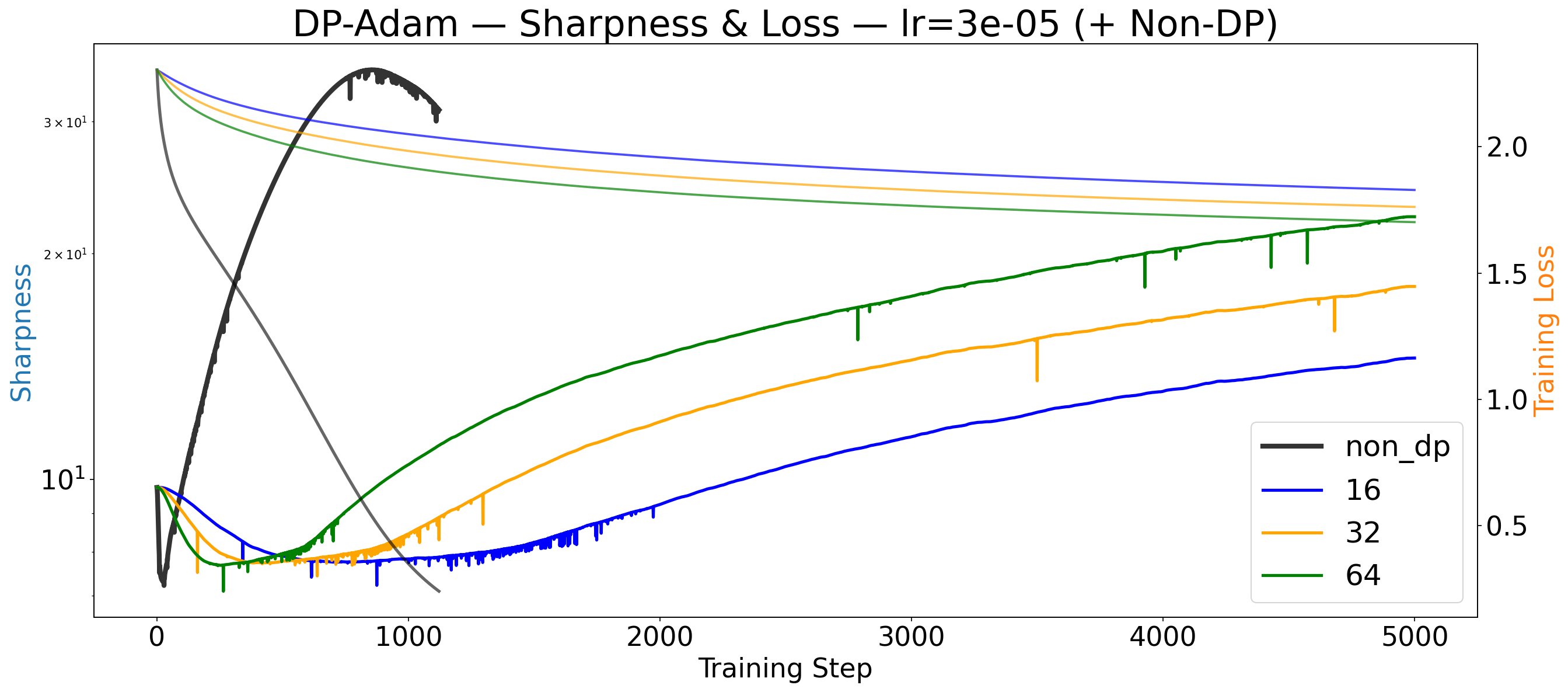}
        \label{fig:row3-col2}
    \end{minipage}

     \caption{
    \textbf{DP-Adam sharpness for smaller learning rates.} 
    Sharpness dynamics for DP-Adam under the same experimental setup as previous figures, shown for learning rates $\eta \in \{3\times10^{-5}, 10^{-4}, 3\times10^{-4}\}$. 
    Panels (a–c) show preconditioned sharpness with the corresponding theoretical stability threshold, while panels (d–f) show raw sharpness. 
    All plots include the non-DP baseline and DP runs for $\varepsilon \in \{16, 32, 64\}$, with training losses overlaid on a secondary axis. 
    Progressive sharpening is observed across all $(\eta, \varepsilon)$ pairs, with higher $\varepsilon$ consistently producing higher sharpness. 
    No configuration reaches the non-DP stability threshold, and clear flattening is only observed at larger learning rates, while smaller $\eta$ exhibit slow convergence with largely monotonic losses.
    }

    \label{fig:dp-adam-all-sharpness}
\end{figure*}

\end{document}